\documentclass[lettersize,journal]{IEEEtran}
\usepackage{amsmath,amssymb,amsfonts}
\usepackage{algorithmic}
\usepackage{algorithm}
\usepackage{array}
\usepackage[caption=false,font=normalsize,labelfont=sf,textfont=sf]{subfig}
\usepackage{textcomp}
\usepackage{stfloats}
\usepackage{url}
\usepackage{verbatim}
\usepackage{graphicx}
\usepackage{cite}
\usepackage{mathtools}        
\usepackage{float} 
\usepackage{booktabs,makecell,multirow}
\newcommand{\cmark}{\checkmark}
\newcommand{\xmark}{\texttimes}

\graphicspath{{Fig/}}

\begin{document}

\title{ZoomSpec: A Physics-Guided Coarse-to-Fine Framework for Wideband Spectrum Sensing}

\author{Zhentao~Yang,~\IEEEmembership{Graduate Student Member,~IEEE,}
        Yixiang~Luomei,~\IEEEmembership{Member,~IEEE,}
        Zhuoyang~Liu,~\IEEEmembership{Graduate Student Member,~IEEE,}
        Zhenyu~Liu,
        Feng~Xu,~\IEEEmembership{Senior Member,~IEEE}%
\thanks{This work was supported in part by the National Key Research and Development Program of China under Grant 2024YFF0505503, and in part by the National Natural Science Foundation of China under Grant W2411057. \emph{(Corresponding authors: Yixiang Luomei and Feng Xu.)}}%
\thanks{The authors are with the Key Laboratory for Information Science of Electromagnetic Waves (MoE), Fudan University, Shanghai 200433, China (e-mail: 24110720175@m.fudan.edu.cn; lmyx@fudan.edu.cn; 20110720062@fudan.edu.cn; 24210720229@m.fudan.edu.cn; fengxu@fudan.edu.cn).}%
}

\maketitle

\begin{abstract}
Wideband spectrum sensing for low-altitude monitoring is critical yet challenging due to the coexistence of heterogeneous protocols, large bandwidths, and non-stationary 
Signal-to-Noise Ratios (SNR). Existing data-driven approaches often treat spectrograms directly as natural images, suffering from a fundamental domain mismatch: they neglect the intrinsic 
time-frequency resolution constraints and spectral leakage, leading to poor visibility for narrowband emissions. To address these limitations, this paper proposes ZoomSpec, 
a physics-guided coarse-to-fine framework that fundamentally restructures the sensing pipeline by integrating signal processing priors with deep learning. Specifically, 
we first introduce a Log-Space Short-Time Fourier Transform (LS-STFT) to overcome the geometric bottleneck of linear spectrograms, effectively sharpening narrowband structures while 
maintaining constant relative resolution. A lightweight Coarse Proposal Net (CPN) is then employed to rapidly screen the full band. 
Crucially, to bridge the gap between coarse detection and fine recognition, we design an Adaptive Heterodyne Low-Pass (AHLP) module. 
Unlike standard neural layers, AHLP functions as a physics-guided signal processing operator that executes center-frequency aligning, bandwidth-matched filtering, and safe decimation, effectively purifying the signal of out-of-band interference. Finally, a Fine Recognition Net (FRN) fuses the purified time-domain I/Q sequence with spectral magnitude via dual-domain attention to jointly refine temporal boundaries and modulation classification. Extensive evaluations on the SpaceNet real-world dataset demonstrate that ZoomSpec achieves a state-of-the-art 78.1 mAP@0.5:0.95. The proposed method not only surpasses existing leaderboard systems but also exhibits superior stability across diverse modulation bandwidths, validating the efficacy of embedding physical mechanisms into data-driven sensing.
\end{abstract}

\section{Introduction} 
\IEEEPARstart{I}{n} recent years, the rapid proliferation of unmanned systems in low-altitude airspace has led to a dramatic surge in wireless service density, 
making the electromagnetic spectrum highly congested, dynamic, and heterogeneous \cite{rn1}. Traditional spectrum management, relying on static frequency planning 
and fixed access, can no longer guarantee the reliability and safety required for critical low-altitude monitoring and control. Consequently, Cognitive Radio, 
which empowers systems to adapt transmission parameters by sensing the electromagnetic environment in real time, has emerged as a key enabler. However, the low-altitude 
channel imposes unique challenges: large observation bandwidths, high platform mobility, and non-stationary signal-to-noise ratio (SNR) demand sensing techniques that 
are robust to severe interference and capable of multi-protocol coexistence \cite{rn2}. 

Existing research on wideband spectrum sensing can be broadly categorized into conventional model-based approaches and data-driven paradigms. Regarding model-based approaches, detectors typically rely on expert-defined statistics, such as energy 
detection \cite{rn3,rn4} and cyclostationary analysis \cite{rn5,rn6,rn7,rn8}. While theoretically sound under idealized stationary assumptions, these methods exhibit 
limited generalization capability when facing diverse protocols or the structural complexity of modern heterogeneous radio environments \cite{rn9}. 

Conversely, within the data-driven paradigm, Deep Learning (DL) methods have shown significant promise in specific sub-tasks, such as Automatic Modulation Recognition 
(AMR) \cite{rn10,rn11,rn12,rn13,rn14,rn15,rn16,rn17,rn18} and signal presence detection \cite{rn26,rn27,rn28,rn29}. Nevertheless, the majority of these works optimize 
isolated stages of the pipeline, lacking a unified framework that jointly handles detection, temporal localization, bandwidth estimation, and modulation recognition. 

More recently, inspired by advancements in computer vision, several studies have attempted the direct adaptation of object detection architectures, such as YOLO 
\cite{rn19,rn20,rn23,rn24} and DETR \cite{rn21,rn22,rn25}, to spectrum sensing tasks. In these approaches, wideband I/Q signals are transformed into time-frequency 
spectrograms and treated as natural images. Despite leveraging mature vision backbones, this paradigm suffers from a fundamental domain mismatch. Unlike natural images, 
spectrograms are inherently constrained by the Heisenberg uncertainty principle \cite{rn35}: improving time resolution inevitably degrades frequency resolution. 
Under a fixed time-frequency tiling, narrowband signals immersed in wideband noise occupy extremely sparse pixel support. This creates a geometric learning bottleneck: 
following spectral leakage and interpolation, the energy of narrowband emissions is diluted, and their boundaries become ambiguous. Consequently, bounding box regression becomes 
highly sensitive to minor shifts, and visual features alone are insufficient for distinguishing weak signals from transient interference or strictly purifying the signal for 
downstream recognition. 

To address these limitations, we propose ZoomSpec, a coarse-to-fine framework that incorporates a novel "focus-and-purify" mechanism to 
fundamentally restructure the wideband sensing pipeline. The proposed framework operates sequentially as follows: First, to resolve the geometric learning bottleneck at the 
representation level, we introduce the Log-Space Short-Time Fourier Transform (LS-STFT) that performs a non-linear mapping of the frequency axis. This ensures constant 
relative resolution and significantly sharpens narrowband visibility. Based on this enhanced representation, a Coarse Proposal Net (CPN) rapidly scans the full observation 
band to generate candidate regions. Subsequently, to bridge the gap between coarse proposals and fine recognition, the Adaptive Heterodyne Low-Pass (AHLP) module functions as 
a physics-guided operator. It translates the CPN's outputs into executable signal processing actions—specifically heterodyning, bandwidth-matched filtering, and safe 
downsampling—thereby effectively purifying the signal of out-of-band noise. Finally, the purified baseband stream is fed into the Fine Recognition Net (FRN), 
which leverages dual-domain attention to execute robust classification. The main contributions of this work are summarized as follows: 
\begin{enumerate}
    \item \textbf{Physics-Guided Sensing Framework:} We propose ZoomSpec, a unified architecture that overcomes the domain mismatch of conventional vision-based detectors by integrating signal processing priors into the deep learning loop. By coupling coarse spectral proposals with fine-grained signal restoration, the framework achieves robust wideband sensing under complex non-stationary conditions.
    
    \item \textbf{Specialized Operators for Geometric and Physical Constraints:} We develop domain-specific modules to resolve the intrinsic bottlenecks of wideband sensing. Specifically, we propose the LS-STFT to overcome the time-frequency resolution trade-off, ensuring constant relative resolution for narrowband visibility. Furthermore, we design the AHLP module to bridge the gap between feature extraction and signal restoration. Unlike standard convolutional layers that implicitly learn spatial features from spectrogram textures, AHLP functions as an explicit, parameter-free DSP operator. It mathematically recovers the SNR via bandwidth-matched filtering and safe decimation, effectively suppressing adjacent-channel interference (ACI) before fine-grained recognition.
    
    \item \textbf{SOTA Performance with Interpretability:} Extensive evaluations on the SpaceNet dataset \cite{rn36}, comprising 14 real-world signal types, demonstrate that ZoomSpec achieves a state-of-the-art 78.1 mAP@0.5:0.95. The proposed method surpasses top leaderboard systems and exhibits superior localization accuracy at high IoU thresholds, validating the efficacy of embedding physical mechanisms into data-driven models.
\end{enumerate}

\section{Signal Model and Problem Formulation}
\label{sec:signal-model}

\subsection{Wideband Signal Model in Low-Altitude Scenarios}
We consider wideband complex baseband observations sampled at rate $F_s$ over a finite window of $N_s$ samples. In dense low-altitude environments, the received waveform is a superposition of $K$ dominant heterogeneous emitters, often overlapping in time and frequency and affected by air-to-ground impairments including residual carrier offsets, Doppler-induced drifts, multipath fading, and adjacent-channel leakage \cite{rn37}. We model the received sequence as
\begin{equation}
\label{eq:rx_rewrite}
\begin{aligned}
r[n]
&=\sum_{k=1}^{K}
e^{\mathrm{j}\left(2\pi \Delta f_k \frac{n}{F_s}+\theta_{k,0}+\theta_k[n]\right)}
\sum_{p=0}^{P_k-1} h_{k,p}[n]\; s_k[n-d_{k,p}] \\
&\quad +\, i[n] + w[n], \qquad n=0,\ldots,N_s-1,
\end{aligned}
\end{equation}
where $s_k[n]$ is the $k$-th transmitted baseband signal; $\Delta f_k$ is the residual frequency offset; $\theta_{k,0}$ is the initial phase; $h_{k,p}[n]$ and $d_{k,p}$ are the time-varying complex gain and discrete delay of the $p$-th multipath tap with $P_k$ taps; $i[n]$ aggregates residual co-channel/adjacent interference beyond the $K$ dominant emitters; and $w[n]$ is additive thermal noise.

To capture oscillator phase noise, the random phase process $\theta_k[n]$ is modeled as a discrete-time Wiener process:
\begin{equation}
\theta_k[n]=\theta_k[n-1]+\nu_k[n], \qquad \nu_k[n]\sim \mathcal{N}(0,\sigma_{\nu,k}^2),
\label{eq:phase_rewrite}
\end{equation}
where $\nu_k[n]$ denotes the Gaussian phase increment.
We write $r[n]\triangleq r_I[n]+\mathrm{j}\,r_Q[n]$ and stack samples into an I/Q matrix
\begin{equation}
\mathbf{r}_{\mathrm{IQ}}\triangleq
\begin{bmatrix}
r_I[0] & r_I[1] & \cdots & r_I[N_s-1]\\
r_Q[0] & r_Q[1] & \cdots & r_Q[N_s-1]
\end{bmatrix}
\in\mathbb{R}^{2\times N_s}.
\label{eq:iq_matrix_rewrite}
\end{equation}

A global $N_s$-point DFT provides a coarse view of spectral occupancy and leakage. With unitary normalization,
\begin{equation}
R_{\mathrm{DFT}}[q]=\frac{1}{\sqrt{N_s}}\sum_{n=0}^{N_s-1} r[n]\; e^{-\mathrm{j}2\pi \frac{q}{N_s}n}, \qquad q=0,\ldots,N_s-1.
\label{eq:dft_rewrite}
\end{equation}

To localize emissions in both time and frequency, we compute the STFT. Let $g[\tau]$ be an analysis window of length $N_w$ and $H$ be the hop size. Denoting the number of frames by $N_f$, the STFT at frame $\ell$ and frequency bin $m$ is
\begin{equation}
\label{eq:stft_rewrite}
\begin{aligned}
X[\ell,m]
&=\sum_{\tau=0}^{N_w-1} r[\tau+\ell H]\; g[\tau]\;
e^{-\mathrm{j}2\pi \frac{m}{M}\tau}, \\
&\qquad \ell=0,\ldots,N_f-1,\quad m=0,\ldots,M-1.
\end{aligned}
\end{equation}
We adopt a log-magnitude rendering
\begin{equation}
S[\ell,m]=\log\!\big(|X[\ell,m]|+\epsilon\big),
\label{eq:logmag_rewrite}
\end{equation}
where $\epsilon>0$ is a numerical constant. Stacking all frames yields $\mathbf{S}\in\mathbb{R}^{N_f\times M}$.

\subsection{Problem Formulation}
Each emission instance is associated with a time support $[t_{s,k},t_{e,k}]$ and an occupied frequency span $[f_{s,k},f_{e,k}]$, equivalently parameterized by center frequency $f_{c,k}=(f_{s,k}+f_{e,k})/2$ and bandwidth $B_k=f_{e,k}-f_{s,k}$. Given the observation $r[n]$ , ZoomSpec aims to detect all active dominant emitters and estimate their parameters.

We infer a set of tuples $\mathcal{O} = \{ \mathbf{o}_1, \dots, \mathbf{o}_{\hat{K}} \}$:
\begin{equation}
\mathbf{o}_k = \left( \hat{t}_{s,k}, \hat{t}_{e,k}, \hat{f}_{c,k}, \hat{B}_k, \hat{\mathcal{C}}_k \right),
\end{equation}
where $\hat{\mathcal{C}}_k$ is the predicted modulation category and $\hat{K}$ is unknown. Parameters are estimated in a coarse-to-fine manner: CPN localizes candidates on LS-STFT and provides coarse band priors; AHLP purifies each candidate by heterodyning, bandwidth-matched low-pass filtering, and safe decimation; FRN refines temporal boundaries, bandwidth, and modulation classification.

\section{Proposed Method: ZoomSpec}
\label{sec:method}

\begin{figure}[!t]
  \centering
  \includegraphics[width=.45\textwidth]{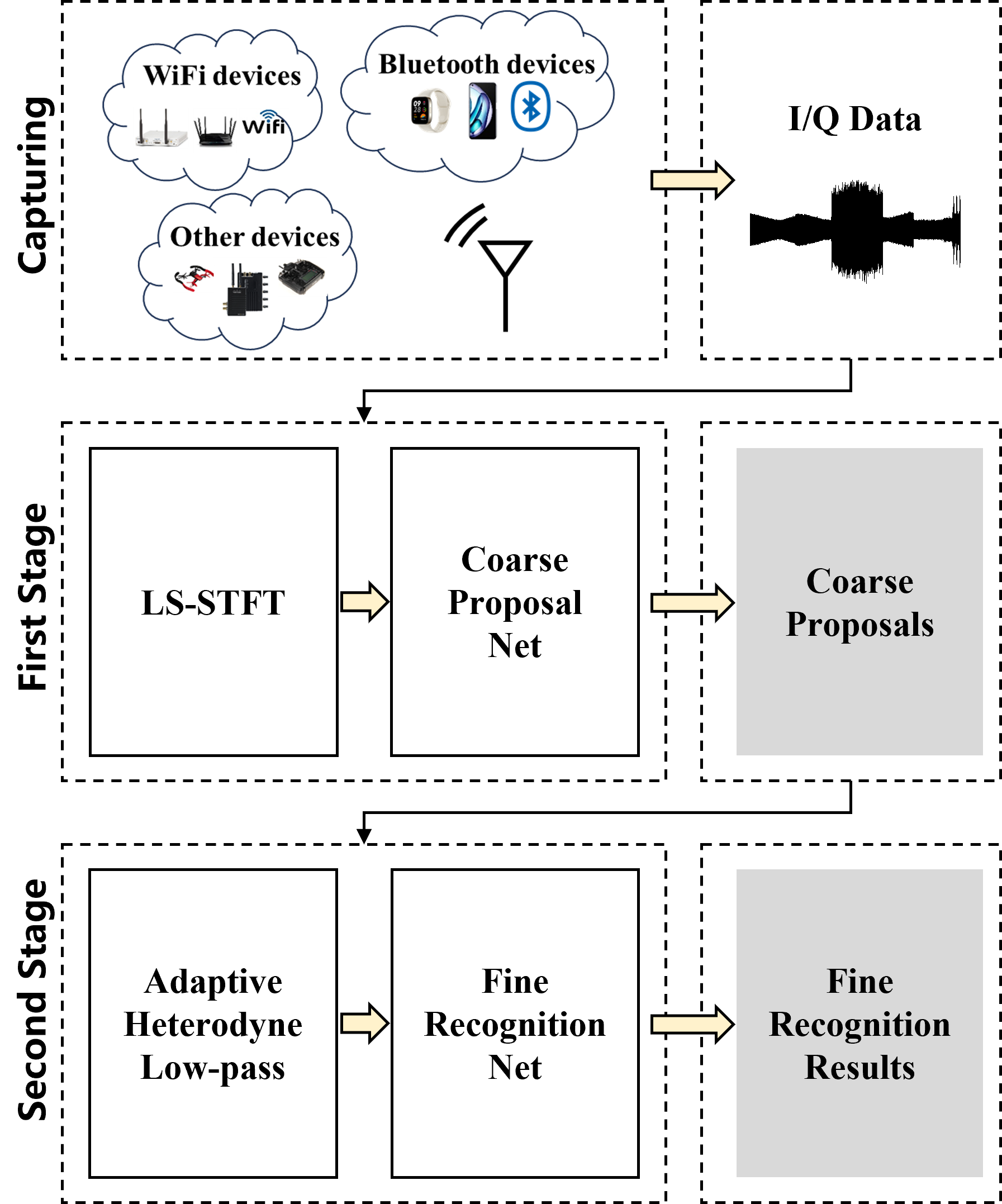}
  \caption{Overview of the proposed ZoomSpec architecture. The system strictly follows a Physics-Guided paradigm: the LS-STFT overcomes the geometric resolution bottleneck, 
  while the AHLP module acts as a physics-guided operator to purify signal candidates before the Dual-Domain FRN.}
  \label{fig:framework}
\end{figure}

This section details ZoomSpec, a coarse-to-fine architecture designed around a focus-and-purify mechanism as shown in Fig.~\ref{fig:framework}. LS-STFT addresses the geometric learning bottleneck by providing near-constant relative resolution over wide bands under a fixed frequency-axis budget. On this representation, CPN screens the full band and outputs a small set of coarse time-frequency proposals. Conditioned on each proposal, AHLP acts as a physics-guided zooming operator that concentrates in-band energy and suppresses out-of-band interference and adjacent leakage, enabling safe decimation (i.e., downsampling that respects the post-filter Nyquist constraint). Finally, FRN fuses purified time-domain I/Q with spectral magnitude via dual-domain attention to jointly refine temporal boundaries, occupied bandwidth, and modulation classification.

\subsection{Log-Space Variable-Resolution Frequency Mapping}
\label{subsec:ls-stft}

Conventional STFT with linear frequency sampling assigns a fixed absolute bin spacing under a fixed canvas size \cite{rn38}. In wideband monitoring, this is inefficient: narrowband emissions occupy very few frequency bins, and their edges are often smeared by leakage or interpolation, impeding precise localization.

To improve narrowband visibility without increasing the total spectrogram size $M$, a periodic log-warping scheme is proposed, which repeats a variable-resolution template across the observation band.

The total bandwidth $B_{\mathrm{obs}}$ is partitioned into $N_{\mathrm{sub}}$ contiguous subbands. The bandwidth of each subband is given by:
\begin{equation}
B_{\mathrm{sub}} \triangleq \frac{B_{\mathrm{obs}}}{N_{\mathrm{sub}}}.
\label{eq:df}
\end{equation}
Accordingly, $M_{\mathrm{sub}}$ frequency points are assigned to each subband, such that the total frequency dimension corresponds to $M = N_{\mathrm{sub}}M_{\mathrm{sub}}$.

A monotonic logarithmic grid is first constructed on a half-interval basis ($i=0,\ldots,M_{\mathrm{sub}}/2-1$). For notational simplicity, a logarithmic step factor $\delta$ is defined as:
\begin{equation}
\delta = \frac{\alpha_2 - \alpha_1}{M_{\mathrm{sub}}/2 - 1},
\label{eq:delta_def}
\end{equation}
where $\alpha_1$ and $\alpha_2$ are hyperparameters controlling the warping curvature. A larger difference $|\alpha_2 - \alpha_1|$ yields a steeper mapping gradient. The base coordinates are then generated by:
\begin{equation}
b_i = \frac{10^{\alpha_1 + i\delta} - 10^{\alpha_1}}{10^{\alpha_2} - 10^{\alpha_1}} \in [0,1].
\label{eq:lsstft-base}
\end{equation}

To ensure continuity at subband boundaries, this base grid is mirrored to form a symmetric, unit-interval template $\{\tilde{b}_j\}_{j=0}^{M_{\mathrm{sub}}-1}$:
\begin{equation}
\tilde{b}_j = 
\begin{cases}
\frac{1}{2}b_j, & 0 \le j < \frac{M_{\mathrm{sub}}}{2}, \\[5pt]
1 - \frac{1}{2}b_{M_{\mathrm{sub}}-1-j}, & \frac{M_{\mathrm{sub}}}{2} \le j < M_{\mathrm{sub}}.
\end{cases}
\label{eq:lsstft-sym}
\end{equation}

This symmetric construction concentrates sampling density at specific regions and ensures smooth transitions between adjacent subbands.

For the $k$-th subband ($k=0,\ldots,N_{\mathrm{sub}}-1$), the warped frequency grid points are defined as:
\begin{equation}
f^{(\log)}_{j,k} = f_{\min} + k B_{\mathrm{sub}} + \tilde{b}_j B_{\mathrm{sub}}, 
\quad j=0,\ldots,M_{\mathrm{sub}}-1.
\label{eq:lsstft-tile}
\end{equation}

Stacking these grids yields the full non-uniform frequency axis.
The warped spectrogram $\widetilde{\mathbf{S}}$ is then obtained by resampling the original linear STFT $X(\ell,m)$ onto $\{f^{(\log)}_{j,k}\}$ via interpolation:
\begin{equation}
\widetilde{X}(\ell,j,k) = \Big[ \mathcal{I}\{ X(\ell, \cdot) \} \Big]_{f=f^{(\log)}_{j,k}}.
\label{eq:lsstft-resample}
\end{equation}

Specifically, bilinear interpolation is employed for the operator $\mathcal{I}(\cdot)$ to balance reconstruction quality and computational efficiency. Finally, the indices are flattened via $\tilde{m}=kM_{\mathrm{sub}}+j$, and log-magnitude scaling is applied:
\begin{equation}
\widetilde{S}(\ell,\tilde{m}) = \log\big( |\widetilde{X}(\ell,j,k)| + \epsilon \big).
\label{eq:lsstft-logmag}
\end{equation}

\begin{figure*}[!t]
\centering
\includegraphics[width=1\textwidth]{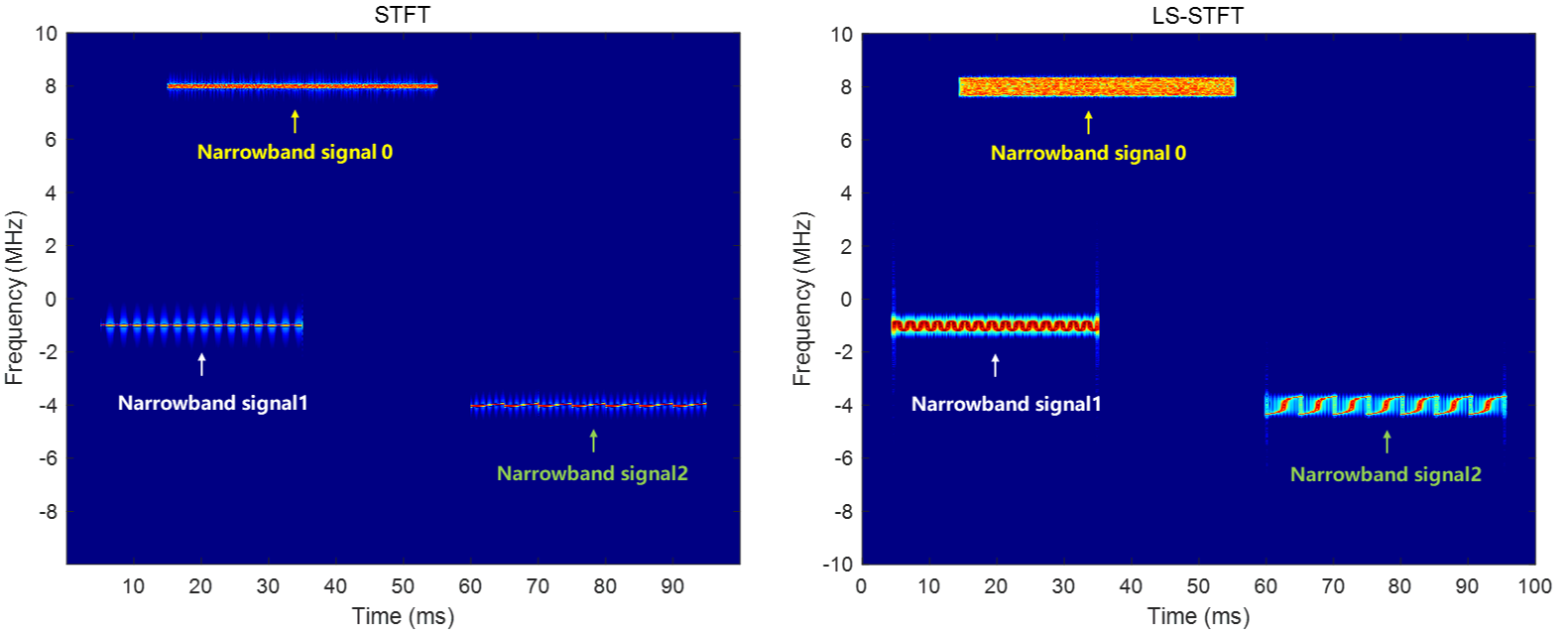} 
\caption{Visual comparison of spectral representations on simulated narrowband signals (Zigbee, LoRa, and NB-FM). While standard STFT suffers from sparse pixel support for narrowband emitters, the proposed LS-STFT significantly expands the visual footprint of signals like LoRa chirps and FM traces through periodic subband warping, enhancing feature prominence for detection.}
\label{fig:sim-compare}
\end{figure*}

The efficacy of this periodic warping is visually validated through a controlled simulation containing three representative narrowband signals: a burst Zigbee signal (Signal 0), a LoRa chirp signal (Signal 1), and a continuous Narrowband FM signal (Signal 2). The comparison between the standard STFT and the proposed LS-STFT is presented in Fig.~\ref{fig:sim-compare}.

In the standard STFT, the narrowband components are constrained to minimal pixel support, resulting in faint features and blurred boundaries. Conversely, in the LS-STFT, the sampling density for these narrowband components is effectively increased. It is observed that the Zigbee burst is rendered with higher contrast, and the geometric details of the signals are significantly expanded. This enhancement provides sharper boundaries and richer texture features for the downstream detection network without increasing the total computational budget.

\subsection{Bandwidth-Aware Coarse Proposals with Adaptive Heterodyne Low-Pass Filtering}

As the first stage of the pipeline, the CPN identifies a small set of time-frequency (T-F) segments likely to contain valid signals. For each candidate, it outputs a proposal vector $\mathbf{b}=([\hat{t}_s,\hat{t}_e],[\hat{f}_s,\hat{f}_e],k_{\mathrm{bw}},\mathrm{conf})$, where $[\hat{t}_s,\hat{t}_e]$ and $[\hat{f}_s,\hat{f}_e]$ denote the predicted time and frequency spans, $k_{\mathrm{bw}}$ is the bandwidth tier, and $\mathrm{conf}$ is the confidence score.
To balance accuracy and latency, we instantiate CPN with \emph{YOLOv11-nano (YOLOv11n)} \cite{rn39}. Benefiting from the LS-STFT representation, which maintains approximately constant relative resolution, narrowband structures gain sufficient pixel density. Thus, a nano-scale detector suffices to provide a strong accuracy-latency trade-off under a fixed compute budget.
The predicted center frequency and bandwidth are derived as $\hat{f}_c=(\hat{f}_s+\hat{f}_e)/2$ and $\hat{B}=\hat{f}_e-\hat{f}_s$. Post-processing employs a frequency-weighted T-F IoU-NMS to suppress redundant proposals.

\begin{figure*}[!t]
  \centering
  \includegraphics[width=.8\textwidth]{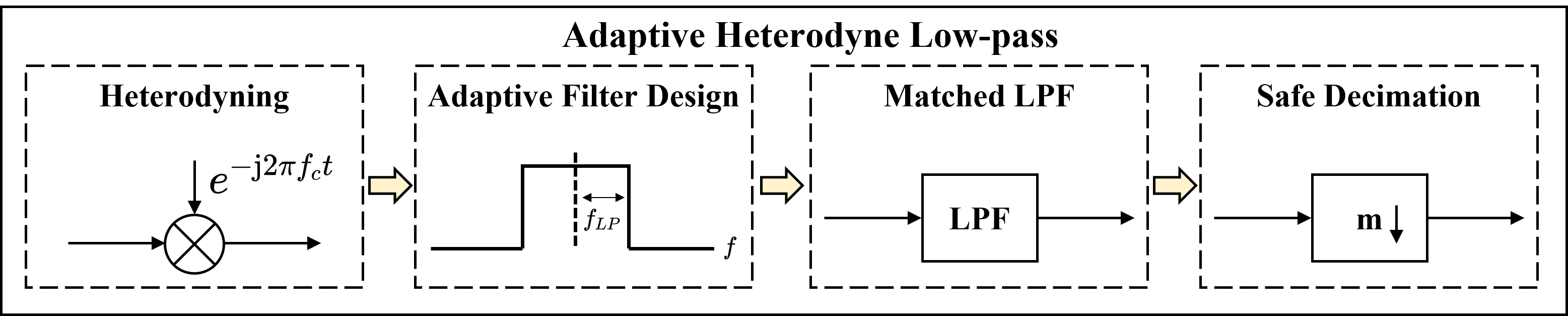} 
  \caption{The AHLP processing chain. Guided by the coarse parameters ($\hat{f}_c$, $\hat{B}$) from the upstream CPN, the module performs baseband translation, bandwidth-matched adaptive filtering, and safe decimation to extract purified I/Q segments.}
  \label{fig:cpn-ahlp}
\end{figure*}

Conditioned on the proposal $\mathbf{b}$, the AHLP module converts these coarse priors into executable signal-processing operators, as illustrated in Fig.~\ref{fig:cpn-ahlp}. The chain consists of three key steps: heterodyning, bandwidth-matched filtering, and safe decimation.

Let the received samples be $r[n]$ at sampling rate $F_s$. We first convert the predicted time span $[\hat{t}_s,\hat{t}_e]$ into sample indices:
\begin{equation}
\hat{n}_s=\lceil \hat{t}_s F_s\rceil,\qquad \hat{n}_e=\lfloor \hat{t}_e F_s\rfloor,\qquad \hat{N}_{\mathrm{seg}}=\hat{n}_e-\hat{n}_s+1,
\label{eq:ahlp-index}
\end{equation}
and define the segment:
\begin{equation} 
r_{\mathrm{seg}}[n]\triangleq r[\hat{n}_s+n] , \qquad n=0,\ldots,\hat{N}_{\mathrm{seg}}-1.
\end{equation}

\subsubsection{Heterodyning}
The segment is shifted to baseband to align the signal of interest with DC:
\begin{equation}
y[n] \;=\; r_{\mathrm{seg}}[n]\; e^{-\mathrm{j}\,2\pi \hat{f}_c (\hat{n}_s+n)/F_s}, \qquad n=0,\ldots,\hat{N}_{\mathrm{seg}}-1.
\label{eq:ahlp-mix}
\end{equation}

\subsubsection{Adaptive Low-Pass Filtering}
A low-pass filter $h_{\mathrm{LP}}[n;\hat{f}_{\mathrm{LP}}]$ is applied to suppress out-of-band interference. The cutoff frequency $\hat{f}_{\mathrm{LP}}$ adapts to the estimated bandwidth $\hat{B}$ and the detection confidence. Since $\hat{B}$ represents the full occupied bandwidth, the one-sided baseband support is approximately $\hat{B}/2$. We set:
\begin{equation}
\hat{f}_{\mathrm{LP}} \;=\; \frac{1}{2}\,\beta(\mathrm{conf})\,\hat{B},
\label{eq:ahlp-cut}
\end{equation}
where $\beta(\mathrm{conf})=1+\kappa\,(1-\mathrm{conf})$ introduces a confidence-dependent guard interval, with $\kappa\in[0.1,0.3]$.
The filtered sequence is given by:
\begin{equation}
z[n] \;=\; (y \ast h_{\mathrm{LP}}[\cdot;\hat{f}_{\mathrm{LP}}])[n].
\label{eq:ahlp-filt}
\end{equation}

\subsubsection{Safe Decimation}
To reduce computational load for the downstream fine recognizer, the signal is decimated. We select the \emph{largest} integer factor $D$ that satisfies the Nyquist condition $F_s/D \ge 2 \hat{f}_{\mathrm{LP}}$:
\begin{equation}
D \;=\; \max\!\left(1,\left\lfloor \frac{F_s}{2\hat{f}_{\mathrm{LP}}} \right\rfloor\right).
\label{eq:ahlp-decim}
\end{equation}
The final purified sequence is:
\begin{equation}
u[m] \;=\; z[mD],\qquad m=0,1,\ldots,\left\lfloor\frac{\hat{N}_{\mathrm{seg}}-1}{D}\right\rfloor.
\label{eq:ahlp-out}
\end{equation}
In practice, $h_{\mathrm{LP}}$ is implemented using a windowed-FIR design with an order proportional to $F_s/\Delta f_{\mathrm{tr}}$, where the transition width $\Delta f_{\mathrm{tr}}\approx \eta \hat{f}_{\mathrm{LP}}$ ($\eta\in(0,1)$).

The overall procedure from proposal generation to purification is summarized in Algorithm~\ref{alg:duaf}.

\begin{algorithm}[t]
\caption{CPN-AHLP}\label{alg:duaf}
\begin{algorithmic}
\STATE \textbf{Input:} samples $r[n]$, sampling rate $F_s$, guard parameter $\kappa$
\STATE \textbf{Output:} purified segments $\{u_i[m]\}$ with metadata
\STATE $\widetilde{\mathbf{S}} \gets \mathrm{LS\mbox{-}STFT}(r[n])$
\STATE $\mathcal{P} \gets \mathrm{CPN}(\widetilde{\mathbf{S}})$; apply frequency-weighted T-F IoU-NMS
\FOR{$i = 1$ \TO $|\mathcal{P}|$}
\STATE extract $\mathbf{b}_i=([\hat{t}_s,\hat{t}_e],[\hat{f}_s,\hat{f}_e],k_{\mathrm{bw}},\mathrm{conf})$
 \STATE $\hat{f}_c \gets (\hat{f}_s+\hat{f}_e)/2$, \ $\hat{B} \gets \hat{f}_e-\hat{f}_s$
 \STATE $\hat{n}_s \gets \lceil \hat{t}_s F_s\rceil$, \ $\hat{n}_e \gets \lfloor \hat{t}_e F_s\rfloor$, \ $\hat{N}_{\mathrm{seg}} \gets \hat{n}_e-\hat{n}_s+1$
 \STATE $r_{\mathrm{seg}}[n] \gets r[\hat{n}_s+n]$ for $n=0,\ldots,\hat{N}_{\mathrm{seg}}-1$
 \STATE $y[n] \gets r_{\mathrm{seg}}[n]\exp\!\big(-\mathrm{j}2\pi \hat{f}_c (\hat{n}_s+n)/F_s\big)$
 \STATE $\beta \gets 1+\kappa(1-\mathrm{conf})$
 \STATE $\hat{f}_{\mathrm{LP}} \gets \tfrac{1}{2}\beta\,\hat{B}$
 \STATE $z[n] \gets (y \ast h_{\mathrm{LP}}[\cdot;\hat{f}_{\mathrm{LP}}])[n]$
 \STATE $D \gets \max\!\left(1,\left\lfloor F_s/(2\hat{f}_{\mathrm{LP}}) \right\rfloor\right)$
 \STATE $u_i[m] \gets z[mD]$
\ENDFOR
\STATE \textbf{return} $\{u_i[m]\}$
\end{algorithmic}
\end{algorithm}

\subsection{Dual-Domain Fine Recognition}\label{sec:frn}

\begin{figure*}[!t]
  \centering
  \includegraphics[width=.9\textwidth]{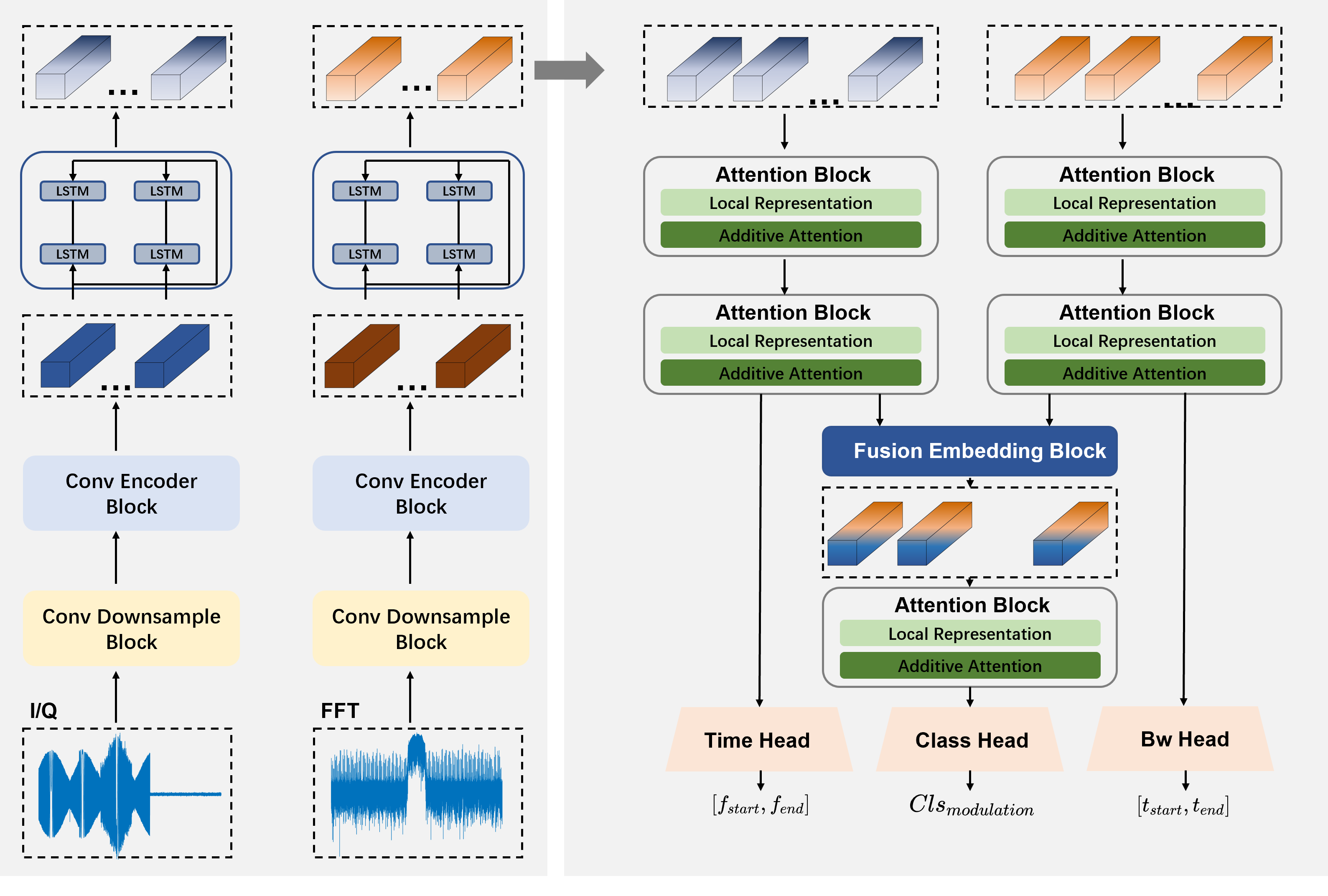}
  \caption{FRN architecture. After AHLP, two streams-time-domain I/Q and FFT magnitude-pass through a 1-D downsampling stem and a shallow conv encoder, followed by a BiLSTM and two local-global additive-attention blocks to yield tokens $\mathbf{A}_{\mathrm{IQ}}$ and $\mathbf{A}_{\mathrm{FFT}}$. A fusion bottleneck mixes channels to form fused tokens $\mathbf{F}$. Heads specialize: Time Head consumes $\mathbf{A}_{\mathrm{IQ}}$; Bw Head consumes $\mathbf{A}_{\mathrm{FFT}}$; Class Head consumes $\mathbf{F}$.}
  \label{fig:duaf}
\end{figure*}

CPN provides coarse proposals and AHLP produces purified baseband segments with suppressed out-of-band energy and improved effective SNR. Building on these inputs, FRN performs fine-grained refinement via dual-domain attention and lightweight task heads as shown in Fig.~\ref{fig:duaf}.

For each proposal, AHLP outputs a purified, baseband, and decimated segment $u_i[m]$:
\begin{equation}
u_i[m] \;=\; z_i[mD_i], \qquad m=0,\ldots,T_i-1,
\label{eq:frn-u}
\end{equation}
where $T_i$ is the segment length after decimation. We define synchronized dual-domain inputs:
\begin{equation}
\mathbf{x}^{(i)}_{\mathrm{IQ}} \triangleq
\begin{bmatrix}
\Re\{u_i\}\\
\Im\{u_i\}
\end{bmatrix}
\in \mathbb{R}^{2\times T_i},
\label{eq:frn-iq}
\end{equation}
and the frequency-domain stream uses magnitude spectrum
\begin{equation}
\mathbf{x}^{(i)}_{\mathrm{FFT}} \triangleq \big|\mathcal{F}\{u_i\}\big|
\in \mathbb{R}^{1\times T_i},
\label{eq:frn-fft}
\end{equation}
where $\mathcal{F}\{\cdot\}$ denotes a 1-D Fourier transform producing a length-$T_i$ vector (full spectrum magnitude is used for simplicity). For brevity, we omit the superscript $i$ when clear and write $T\equiv T_i$.

\subsubsection{Dual-Domain Encoder with Local-Global Additive Attention}\label{sec:frn-encoder}

FRN leverages complementary cues from time and frequency domains. Time-domain features capture instantaneous phase/amplitude dynamics and burst boundaries, while frequency-domain features encode occupied-span geometry and spectral structure under linear-time global modeling. The encoder comprises per-domain stems, bidirectional temporal context encoding, domain-wise local-global additive attention \cite{rn41}, and a lightweight cross-domain fusion bottleneck.

Each domain uses a 1-D stem that downsamples length via a depthwise-separable convolution with stride $s$, followed by channel expansion via a pointwise $1{\times}1$ convolution. After embedding to width $C$, both streams become token sequences
\begin{equation}
\mathbf{H}_{\mathrm{IQ}} \in \mathbb{R}^{C\times L}, \qquad
\mathbf{H}_{\mathrm{FFT}} \in \mathbb{R}^{C\times L},
\label{eq:frn-stem}
\end{equation}
where $L$ is the stem output length (shared across domains).

To aggregate short-range symbol dynamics and long-range envelope periodicity, each branch uses a bidirectional LSTM \cite{rn40}:
\begin{equation}
\widetilde{\mathbf{H}} \;=\; \mathrm{BiLSTM}(\mathbf{H}) \;\in\; \mathbb{R}^{2C\times L}.
\label{eq:frn-bilstm}
\end{equation}

A depthwise 1-D convolution followed by a pointwise convolution with GELU and normalization captures local morphologies without changing length:
\begin{equation}
\mathbf{Z} \;=\; \mathrm{PWConv}\!\big(\mathrm{DWConv1d}(\widetilde{\mathbf{H}})\big) \;\in\; \mathbb{R}^{2C\times L}.
\label{eq:frn-local}
\end{equation}

Let $\mathbf{Z}^\top \in \mathbb{R}^{L\times 2C}$. Queries and keys are
\begin{equation}
\mathbf{Q} \;=\; \mathbf{Z}^\top \mathbf{W}_q, \qquad
\mathbf{K} \;=\; \mathbf{Z}^\top \mathbf{W}_k ,
\label{eq:frn-qk}
\end{equation}
and a learnable global gate $\mathbf{w}_g \in \mathbb{R}^{2C}$ produces weights
\begin{equation}
\boldsymbol{\alpha} \;=\; \mathrm{softmax}(\mathbf{Q}\,\mathbf{w}_g) \;\in\; \mathbb{R}^{L}.
\label{eq:frn-gate}
\end{equation}

The global summary is
\begin{equation}
\mathbf{g} \;=\; \sum_{t=1}^{L} \alpha_t \,\mathbf{Q}_t \;\in\; \mathbb{R}^{2C}.
\label{eq:frn-global}
\end{equation}

Broadcast $\mathbf{g}$ to length $L$, reweight $\mathbf{K}$ elementwise, and apply a linear projection with residual to obtain the block output (with normalization/layer scaling and Dropout/DropPath). Stacking two such blocks per domain yields
\begin{equation}
\mathbf{A}_{\mathrm{IQ}},\; \mathbf{A}_{\mathrm{FFT}} \;\in\; \mathbb{R}^{2C\times L}.
\label{eq:frn-A}
\end{equation}

This additive attention maintains $\mathcal{O}(CL)$ complexity and is robust to spiky/sparse patterns, avoiding the quadratic cost of dot-product self-attention.
\subsubsection{Cross-domain Fusion Bottleneck}
To fuse complementary evidence under tight compute budgets, concatenate and mix channels through a lightweight bottleneck:
\begin{equation}
\mathbf{F}_{\mathrm{cat}}
=\big[\mathbf{A}_{\mathrm{IQ}};\mathbf{A}_{\mathrm{FFT}}\big]
\in \mathbb{R}^{4C\times L},
\label{eq:frn-fcat}
\end{equation}
\begin{equation}
\widehat{\mathbf{F}}
=\mathrm{Norm}\!\big(\mathrm{PWConv}_1(\mathbf{F}_{\mathrm{cat}})\big)
\in \mathbb{R}^{2C\times L},
\label{eq:frn-fuse-mid}
\end{equation}
\begin{equation}
\mathbf{F}
=\mathrm{PWConv}_2\!\big(\mathrm{GELU}(\widehat{\mathbf{F}})\big)
\in \mathbb{R}^{2C\times L}.
\label{eq:frn-fuse}
\end{equation}

Optionally, a final additive-attention block can refine $\mathbf{F}$ for decoding.

\subsubsection{Task Heads and Decoding}\label{sec:frn-heads}
Three lightweight heads share the fused backbone but specialize in interval localization, bandwidth estimation, and modulation classification. Decoding is differentiable and constraint-aware.

\paragraph*{Time head}
On a normalized grid $\xi_t\in[0,1]$, $t=1,\ldots,L$, the head predicts distributions for onset and duration:
\begin{equation}
p_{\mathrm{start}}[t], \qquad p_{\mathrm{dur}}[t].
\label{eq:frn-time-dists}
\end{equation}

Continuous estimates are decoded by expectations:
\begin{equation}
\widehat{t}_{\mathrm{start}} \;=\; \sum_{t=1}^{L} \xi_t \, p_{\mathrm{start}}[t],
\qquad
\widehat{d} \;=\; \sum_{t=1}^{L} \xi_t \, p_{\mathrm{dur}}[t],
\label{eq:frn-time-exp}
\end{equation}
\begin{equation}
\widehat{t}_{\mathrm{end}} \;=\; \min\!\big(1-\varepsilon,\ \widehat{t}_{\mathrm{start}}+\widehat{d}\big), \quad \varepsilon\in(0,10^{-3}].
\label{eq:frn-time-end}
\end{equation}

This decoding avoids argmax/thresholding and enforces valid intervals.

\paragraph*{Bandwidth head}
Since AHLP heterodynes candidates to baseband, occupied span is estimated from frequency-domain tokens. The head outputs a distribution $p_{\mathrm{bw}}[t]$ over a normalized bandwidth grid $\{\xi_t\}_{t=1}^L$, and the bandwidth is decoded by expectation:
\begin{equation}
\widehat{B} \;=\; \sum_{t=1}^{L} \xi_t \, p_{\mathrm{bw}}[t].
\label{eq:frn-bw-exp}
\end{equation}

\paragraph*{Modulation head}
A linear classifier with softmax on $\mathbf{F}$ predicts $\mathbf{p}_{\mathrm{cls}}\in\Delta^{N_{\mathrm{cls}}-1}$, where $N_{\mathrm{cls}}$ is the number of modulation classes.

\section{Experiments}\label{sec:dataset}

\subsection{Dataset}
\begin{table}[t]
\centering
\caption{SpaceNet dataset acquisition settings and split.}
\label{tab:spacenet-settings}
\renewcommand{\arraystretch}{1.15}
\begin{tabular}{lcc}
\toprule
\textbf{Dataset} & \multicolumn{2}{c}{\textbf{SpaceNet}}\\
\midrule
Modulation families & \multicolumn{2}{c}{14} \\
\midrule
\multirow{2}{*}{\makecell[l]{Sampling rate (MHz)}} 
  & \textbf{Train} & 5, 20, 30, 40, 50, 80 \\
\cmidrule(lr){2-3}
  & \textbf{Test}  & 20, 30, 40, 50, 80 \\
\midrule
\multirow{2}{*}{\makecell[l]{Duration (ms)}} 
  & \textbf{Train} & 20, 40, 60, 80, 100, 150 \\
\cmidrule(lr){2-3}
  & \textbf{Test}  & 20, 40, 60, 80, 100, 200 \\
\midrule
\multirow{2}{*}{Signals per file}
  & \textbf{Train} & 1-8 \\
\cmidrule(lr){2-3}
  & \textbf{Test}  & 6-10 \\
\bottomrule
\end{tabular}
\end{table}

\begin{figure}[h]
\centering
\includegraphics[width=\columnwidth]{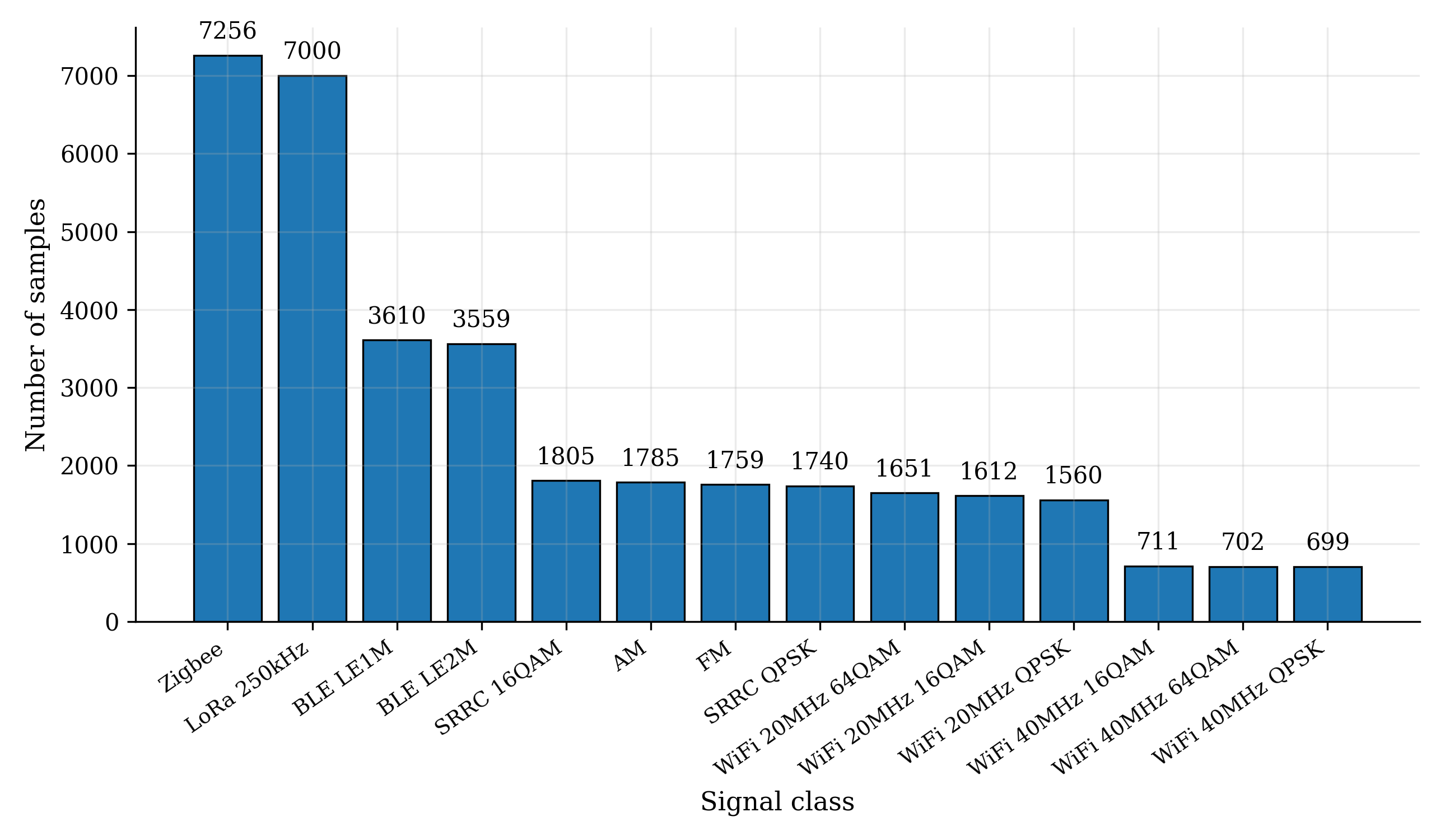}
\caption{Class distribution of SpaceNet dataset.}
\label{fig:spacenet-classes}
\end{figure}
SpaceNet dataset is jointly curated by the Institute of Space Internet of Fudan University and the Shanghai Radio Monitoring Station, and serves as the official dataset of the 2025 ``AI+Radio'' Challenge\cite{rn44}. 
All experiments are conducted on the SpaceNet public real-world benchmark, which covers the entire 2.4-2.4835~GHz ISM band. 
Measurements were collected in representative low-altitude scenarios spanning urban, suburban, indoor, and open areas, and then composed into multi-signal scenes under a single-signal acquisition plus controlled composition protocol that allows controlled overlaps in both time and frequency. 
The corpus combines field captures with an equal proportion of high-fidelity MATLAB simulations, yielding approximately 10{,}000 labeled samples across 14 modulation families. 
Acquisition settings are summarized in Table~\ref{tab:spacenet-settings}, and per-class counts are shown in Fig.~\ref{fig:spacenet-classes}.

Each sample provides a complex I/Q time series in binary format, accompanied by a .json annotation specifying the center frequency, occupied bandwidth, class label, and active time interval. 
We follow the official train/test split, where the test set is intentionally denser in terms of concurrent emitters and overlaps so as to better reflect low-altitude electromagnetic conditions. 
Unless otherwise stated, all reported results concern spectrum-occupancy detection, bandwidth estimation, and modulation recognition on the official SpaceNet test split. 

\subsection{Compared Methods}

To strictly assess the effectiveness of the proposed ZoomSpec framework, we establish a robust baseline comparison strategy. It is worth noting that traditional spectrum sensing methods 
(e.g., energy detection, cyclostationary feature detection) are omitted from this comparison. Preliminary experiments indicated that these model-based approaches fail to handle the 
high-density overlaps and heterogeneous signal types present in the SpaceNet dataset, resulting in negligible detection mAP. 

Instead, we focus our comparison on Deep Learning-based Object Detectors, which have emerged as the dominant paradigm in the associated ``AI+Radio'' Challenge \cite{rn44}. 
Observations from the challenge leaderboard reveal that top-performing solutions are predominantly variants of one-stage detectors (YOLO family) and Transformer-based detectors (DETR family). 
To ensure a scientifically fair and reproducible comparison, rather than replicating the ad-hoc ensembles or contest-specific engineering heuristics used by individual competition teams, 
we adopt the standard, official implementations of the three most representative architectures. This approach isolates the algorithmic contributions from implementation tricks. 
Specifically, we re-train the following baselines on SpaceNet dataset with identical preprocessing, input resolution, and learning schedules to ZoomSpec:

\textbf{YOLO11 \cite{rn39}:} A one-stage anchor-free detector with a decoupled head, re-parameterizable convolutions, and lightweight attention, followed by NMS at inference. This family represents a high-throughput and deployment-mature real-time paradigm.

\textbf{D-FINE \cite{rn42}:} A DETR-style refinement model that enhances localization via fine-grained distributional box refinement and layer-wise self-distillation while maintaining end-to-end training with Hungarian matching. It primarily improves box quality at high IoU with modest overhead.

\textbf{RF-DETR \cite{rn43}:} A lightweight DETR variant that reduces decoder burden through sparse queries and simplified cross-scale interaction while keeping the end-to-end paradigm and Hungarian matching. It targets stable detection under a tighter compute budget.

It is important to acknowledge that some top-ranking teams on the challenge leaderboard achieve scores slightly higher than our baseline reproductions (e.g., reaching 76-77 mAP with 
similar backbones)\cite{rn45}. Analysis reveals that these entries heavily rely on competition-specific engineering tricks, such as multi-model ensembles, multi-scale testing, and aggressive 
Test-Time Augmentation (TTA). While effective for boosting leaderboard rankings, these techniques obscure the intrinsic contribution of the model architecture and drastically increase 
inference latency. 

To ensure a rigorous scientific evaluation, we exclude such heuristic tricks. All reported results, including our proposed ZoomSpec and the baselines, are evaluated in a 
single-model, single-scale inference mode without TTA. Under this strictly fair comparison, ZoomSpec (78.1 mAP) not only outperforms the standard baselines by a large margin but also surpasses 
the best ensemble-based result on the leaderboard (77.52 mAP)~\cite{rn45}, highlighting the superiority of the proposed physics-guided architecture.

\subsection{Model Implementation}
\label{subsec:implementation}

All models are implemented in Python 3.9 using PyTorch 2.1 with CUDA 12.1 acceleration on a single NVIDIA GeForce RTX 4090 GPU. 
Training employs the AdamW optimizer with an initial learning rate of \(5\times10^{-4}\) and weight decay of \(1\times10^{-4}\). The batch size is set to \(16\), spanning a maximum of \(100\) epochs with early stopping triggered after \(10\) epochs of no validation improvement. 
Unless otherwise noted, input resolution is fixed at $640 \times 640$ for all methods.

Unlike purely data-driven approaches, the architectural hyperparameters of ZoomSpec are grounded in the physical characteristics of the target spectrum, as summarized in Table~\ref{tab:hyperparams}.

For the LS-STFT module, we explicitly set the subband bandwidth $B_{\mathrm{sub}}=1$~MHz. This design choice is twofold: first, typical narrowband emissions possess bandwidths strictly less than $1$~MHz, ensuring they are fully encapsulated within a single warping period for detail amplification; second, the other waveforms typically exhibit bandwidths that are integer multiples of $1$~MHz, so aligning $B_{\mathrm{sub}}$ to $1$~MHz ensures that their relative spectral occupancy remains consistent across the warped axis. 
Regarding the warping curvature, we empirically select $\alpha_1=1$ and $\alpha_2=4$. This specific range is critical: a smaller $\alpha_2$ fails to provide sufficient magnification for fine-grained features, whereas an excessively large $\alpha_2$ over-concentrates sampling density at the subband center, thereby suppressing the discriminability between narrowband signals of slightly different bandwidths.

For the AHLP module, the filter $h_{\mathrm{LP}}$ utilizes a Hamming window to balance main-lobe width and stopband attenuation. The guard parameter is set to $\kappa=0.2$ to safely accommodate proposal errors. Crucially, to ensure real-time performance, the filtering is implemented as a vectorized frequency-domain multiplication on the GPU, allowing concurrent processing of all proposals within a unified tensor batch.

\begin{table*}[t]
\centering
\caption{Implementation details and physical justifications for hyperparameters.}
\label{tab:hyperparams}
\renewcommand{\arraystretch}{1.2}
\setlength{\tabcolsep}{4pt}
\begin{tabular}{l c c l}
\toprule
\textbf{Module} & \textbf{Parameter} & \textbf{Value} & \textbf{Physical/Empirical Justification} \\
\midrule
\multirow{3}{*}{LS-STFT}
 & $B_{\mathrm{sub}}$ & 1 MHz & Matches typical narrowband width and integer channel align. \\
 & $\alpha_1, \alpha_2$ & 1, 4 & Prevents over-concentration while magnifying details. \\
\midrule
\multirow{2}{*}{AHLP} 
 & $\kappa$ & 0.2 & Moderate bandwidth expansion for safety. \\
 & $\eta$ & 0.1 & Standard transition width (10\%) for FIR design. \\
\midrule
\multirow{3}{*}{Training} 
 & Optimizer & AdamW & Better generalization stability. \\
 & Learning Rate & $5\times10^{-4}$ & Standard for Transformer-based architectures. \\
 & Batch Size & 16 & Optimized for GPU memory utilization. \\
\bottomrule
\end{tabular}
\end{table*}

\subsection{Evaluations}

Unless otherwise specified, detection performance is evaluated using mAP@0.5:0.95 and precision/recall at IoU~$=0.5$ on SpaceNet dataset. 
We first compare overall performance, then analyze robustness under varying IoU thresholds and across different modulation types.

\subsubsection{Visual Analysis of Spectral Representations}

\begin{figure*}[!t]
  \centering
  \includegraphics[width=0.9\textwidth]{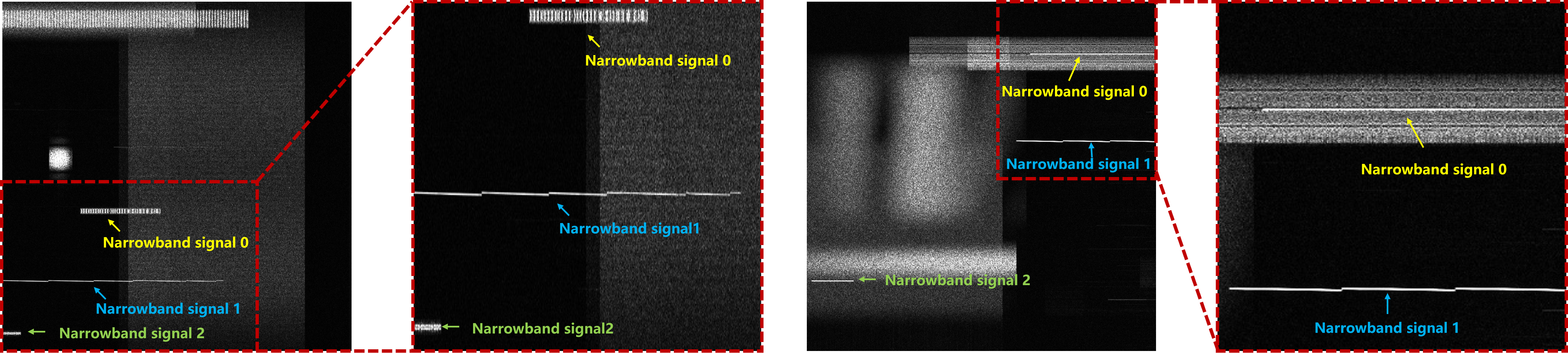}
  \caption{Visualization of the standard STFT spectrogram. Due to the linear frequency sampling, narrowband emissions (highlighted in the zoomed insets) occupy very few pixels on the frequency axis, resulting in weak contrast and blurred boundaries that impede small-object detection.}
  \label{fig:stft-linear}
\end{figure*}

The efficacy of different spectral representations is qualitatively analyzed using real-world samples from the SpaceNet dataset. 
The standard STFT spectrogram is presented in Fig.~\ref{fig:stft-linear}. 
It is observed that due to linear frequency sampling, narrowband emissions are constrained to occupy minimal pixel support along the frequency axis. 
Consequently, weak contrast and indistinct boundaries are exhibited, presenting a significant challenge for small-object detection.

\begin{figure*}[!t] 
  \centering
  \includegraphics[width=0.9\textwidth]{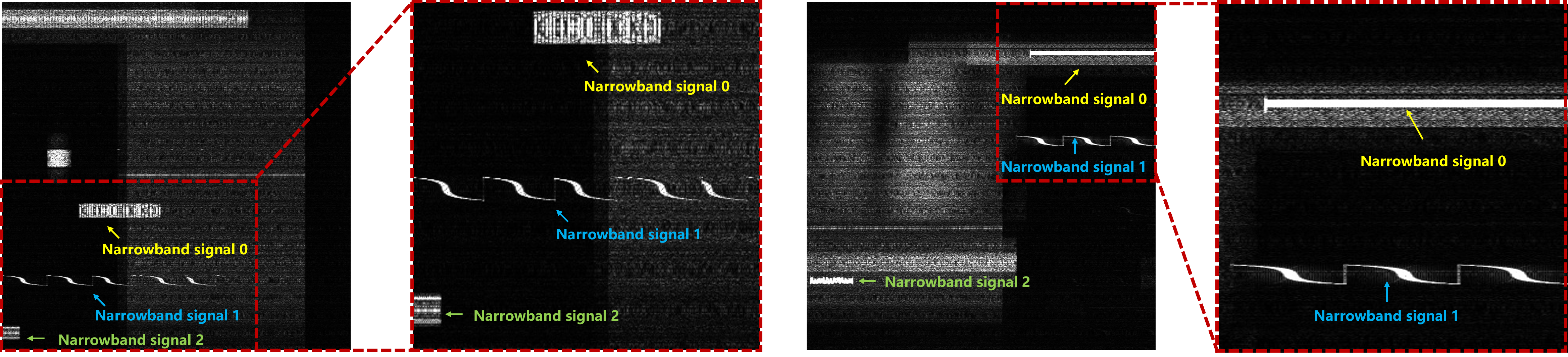}
  \caption{Visualization of the proposed LS-STFT spectrogram under the same frequency budget $M$. The periodic subband warping effectively increases the sampling density for narrowband components, significantly enhancing their visual prominence and sharpness for the detection network.}
  \label{fig:stft-log}
\end{figure*}

Conversely, the proposed LS-STFT spectrogram is illustrated in Fig.~\ref{fig:stft-log}, computed under an identical frequency budget $M$. 
Through periodic subband warping, the sampling density allocated to narrowband components is effectively increased. 
The visual prominence and structural sharpness of these signals are significantly enhanced, providing a more discriminative representation for the detection network.

\subsubsection{Overall Detection Results on SpaceNet dataset}

Table~\ref{tab:main-results} reports the main detection results on SpaceNet dataset in terms of mAP@0.5:0.95 and precision/recall at IoU equal to 0.5. 
On the linear-frequency STFT input, the three detector families already span a broad operating range. 
YOLO11$_\text{STFT}$ offers the lowest latency but only reaches 50.3 mAP@0.5:0.95 with a relatively conservative recall of 0.62. 
RF-DETR$_\text{STFT}$ slightly improves mAP to 53.4 and recall to 0.66, reflecting the benefit of end-to-end matching and global attention, while D-FINE$_\text{STFT}$ further pushes mAP to 60.6 and recall to 0.73 by refining box quality at high IoU. 
This hierarchy suggests that SpaceNet dataset is not trivial for standard detectors, especially when narrow-band and wide-band signals must be handled within a single model.

Replacing the linear-frequency STFT with the proposed log-space mapping LS-STFT yields consistent and substantial gains across all three detector families. 
On YOLO11, mAP@0.5:0.95 improves from 50.3 to 62.7, which corresponds to a relative gain of about $25\%$, and both precision and recall increase from 0.76 and 0.62 to 0.83 and 0.76. 
On RF-DETR, mAP@0.5:0.95 rises from 53.4 to 69.9, accompanied by precision and recall gains from 0.78 and 0.66 to 0.86 and 0.81. 
On D-FINE, mAP@0.5:0.95 improves from 60.6 to 74.3, with precision and recall moving from 0.82 and 0.73 to 0.88 and 0.83. 
The larger margins observed for the DETR family indicate that LS-STFT particularly stabilizes localization at higher IoU thresholds, where precise alignment of spectral boundaries is critical. 
Overall, these improvements confirm that constant relative resolution and sharpened narrow-band structures benefit both one-stage and transformer-based detectors, without changing their architectures or loss functions.

On top of LS-STFT, the proposed method achieves the best overall accuracy at 78.1 mAP@0.5:0.95 with a balanced precision and recall of 0.90 and 0.86. 
Compared with the strongest LS-STFT baseline D-FINE, this corresponds to a further gain of 3.8 mAP points and noticeable improvements in both precision and recall. 
This suggests that physics-guided AHLP purification and dual-domain fusion do more than simply provide another front-end. 
By suppressing spectral leakage and interference before detection, the model reduces false alarms while still recovering weak true positives, thus moving the operating point closer to the ideal upper right corner of the precision–recall trade-off.

\begin{table}[t]
\centering
\caption{Detection performance comparison on SpaceNet dataset. Integrating the proposed LS-STFT consistently improves baseline detectors, while the full ZoomSpec framework achieves state-of-the-art results.}
\label{tab:main-results}
\renewcommand{\arraystretch}{1.2}
\setlength{\tabcolsep}{10pt}
\begin{tabular}{lccc}
\toprule
\textbf{Method} & \textbf{mAP\textsubscript{50:95}} & \textbf{Precision} & \textbf{Recall} \\
\midrule
YOLO11\textsubscript{STFT}    & 50.3 & 0.76 & 0.62 \\
YOLO11\textsubscript{LS-STFT} & 62.7 & 0.83 & 0.76 \\
\addlinespace
RF-DETR\textsubscript{STFT}   & 53.4 & 0.78 & 0.66 \\
RF-DETR\textsubscript{LS-STFT}& 69.9 & 0.86 & 0.81 \\
\addlinespace
D-FINE\textsubscript{STFT}    & 60.6 & 0.82 & 0.73 \\
D-FINE\textsubscript{LS-STFT} & 74.3 & 0.88 & 0.83 \\
\midrule
\textbf{ZoomSpec (Ours)}      & \textbf{78.1} & \textbf{0.90} & \textbf{0.86} \\
\bottomrule
\end{tabular}
\end{table}

\begin{figure}[!t]
  \centering
  \includegraphics[width=\columnwidth]{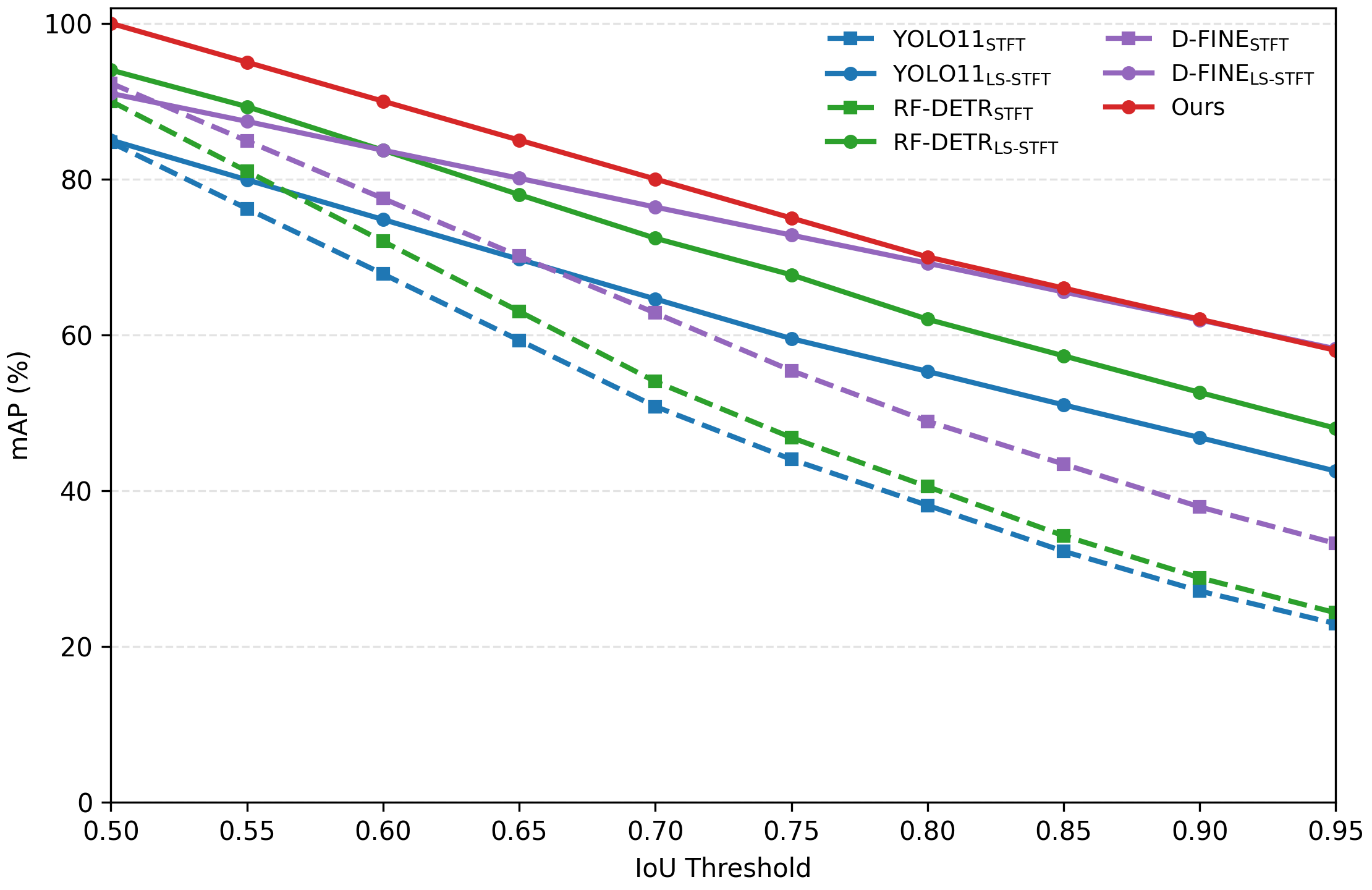}
  \caption{mAP versus IoU threshold on SpaceNet dataset. Color indicates detector family; solid lines are LS-STFT and dashed lines are STFT.}
  \label{fig:ap-iou}
\end{figure}

\subsubsection{Robustness across IoU thresholds}

mAP is plotted as a function of the IoU threshold on the SpaceNet dataset in Figure~\ref{fig:ap-iou}.

Curves with the same color belong to the same detector family, where solid lines use LS-STFT and dashed lines use the linear-frequency STFT. 
This view makes the localization behaviour at different overlap requirements explicit and complements the single-number summary in Table~\ref{tab:main-results}.

For all three detector families, the dashed curves corresponding to the linear-frequency STFT show a relatively sharp decay once the IoU threshold exceeds about 0.75. 
This behaviour is typical when the representation does not provide enough resolution or contrast around spectral edges: boxes can roughly cover active bands but tend to drift at their boundaries, which is heavily penalized at high IoU. 
The solid LS-STFT curves are uniformly higher and noticeably flatter, especially for the DETR family. 
The gap between solid and dashed curves widens as the IoU threshold approaches 0.9 and 0.95, which indicates that the log-space mapping improves not only detection of whether a signal exists, but also the geometric accuracy of the predicted bandwidth and center frequency.

Within each family, LS-STFT also changes the relative ordering of methods. 
For YOLO11, the curve with LS-STFT extends the useful operating range from roughly 0.8 to 0.9 IoU before a steep drop, making the one-stage detector much more competitive when strict localization is required. 
For RF-DETR and D-FINE, the LS-STFT curves stay above 0.6 mAP even at IoU equal to 0.9, whereas their STFT counterparts have already collapsed. 
The proposed method maintains the highest curve over the entire IoU range. 
At low thresholds it inherits the strong recall of LS-STFT-enhanced detectors, and at high thresholds it preserves the best localization accuracy, reflecting the combined effect of AHLP purification and dual-domain fusion. 
The area under each curve closely matches the corresponding mAP@0.5:0.95 in Table~\ref{tab:main-results}, which validates that the curves capture consistent ranking and margin across different localization regimes.

\subsubsection{Robustness across modulation types}

Table~\ref{tab:cls-robustness} further breaks down mAP@0.5:0.95 by modulation type. 
Across all compared methods, replacing the linear-frequency STFT with LS-STFT consistently improves per-class detection accuracy. 
For WiFi waveforms, LS-STFT mainly stabilizes performance across different bandwidths and constellations. 
When moving from 20~MHz to 40~MHz or from QPSK to 64-QAM, the mAP improvement is typically between $5$ and $15$ points, which indicates that the proposed frequency remapping alleviates the resolution imbalance between narrowband and wider-band OFDM spectra. 
On narrowband and bursty non-OFDM signals, the performance gain is even more pronounced. 
BLE, Zigbee, and LoRa all exhibit clear improvements when switching from STFT to LS-STFT; for example, BLE~LE2M under YOLO11 increases from $43.8$ to $59.7$ mAP, and Zigbee increases from $35.7$ to $58.7$ mAP. 
These results show that sharpening narrow spectral lines and transient segments is crucial for reliable detection.
\begin{figure*}[!t]
    \centering
    \includegraphics[width=\textwidth]{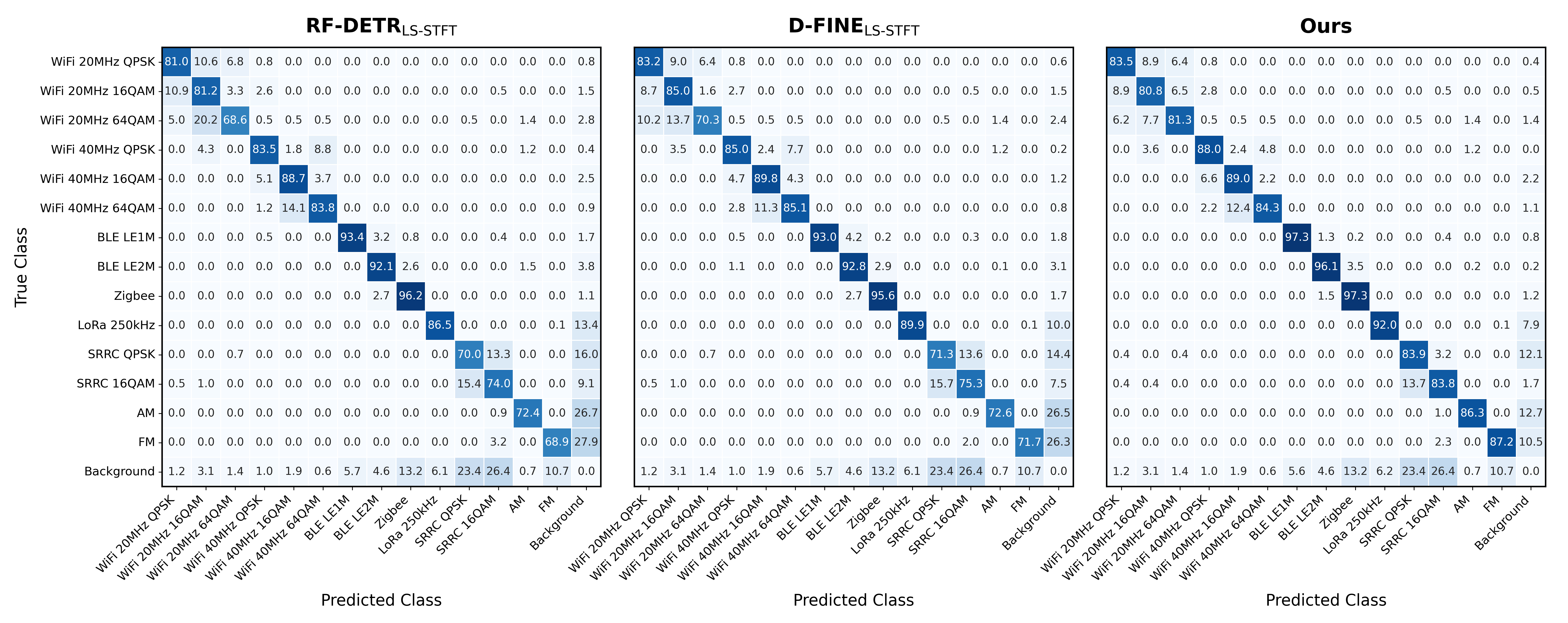}
    \caption{Per-class confusion matrices comparison on SpaceNet dataset. 
    From left to right: RF-DETR$_{\text{LS-STFT}}$, D-FINE$_{\text{LS-STFT}}$, and the proposed method (Ours). 
    The diagonal elements represent the classification recall, while off-diagonal elements indicate misclassification rates. 
    Our method achieves the highest diagonal density across all categories, showing significant robustness in distinguishing narrowband signals (e.g., Zigbee, LoRa) and suppressing background confusion compared to the baselines.}
    \label{fig:cm-compare}
\end{figure*}

On top of LS-STFT, the proposed method achieves the best mAP on all modulation categories. 
Compared with the strongest LS-STFT baseline, our detector improves mAP by about $5$ to $10$ points on most non-WiFi classes, for example BLE~LE2M from $78.7$ to $86.8$, Zigbee from $79.8$ to $87.1$, and AM from $71.5$ to $83.6$, and it still delivers consistent gains on high-SNR WiFi traffic. 
These trends indicate that physics-guided AHLP purification and dual-domain feature fusion not only improve overall mAP, but also enhance robustness to heterogeneous signal structures, ranging from wideband multicarrier OFDM to narrowband single-carrier and analog modulations.

The per-class confusion patterns for D-FINE$_{\text{LS-STFT}}$, RF-DETR$_{\text{LS-STFT}}$, and the proposed method are visualized in Figure~\ref{fig:cm-compare}.
LS-STFT already produces diagonally dominant confusion matrices, yet D-FINE and RF-DETR still exhibit noticeable off-diagonal mass, especially among WiFi modulation orders and between narrowband IoT signals and the background. 
The proposed method further concentrates probability mass on the main diagonal and suppresses cross-modulation errors. 
For BLE, Zigbee, and LoRa, the correct recognition rates all exceed $90\%$, while false alarms into these classes and into the background row are significantly reduced. 
The cleaner background row in our confusion matrix confirms that AHLP effectively removes spurious spectral leakage, allowing the detector to distinguish weak modulated signals from clutter-like interference. 
Taken together, the per-class mAP and confusion-matrix analysis demonstrate that the proposed dual-domain processing pipeline achieves substantially stronger robustness across diverse modulation types than LS-STFT-enhanced baselines.

\begin{table*}[t]
\centering
\caption{Per-class mAP@0.5:0.95 comparison on SpaceNet dataset. The proposed LS-STFT input significantly boosts detection performance for narrowband and weak signals (e.g., Zigbee, LoRa, AM/FM) compared to standard STFT.}
\label{tab:cls-robustness}
\renewcommand{\arraystretch}{1.25}
\setlength{\tabcolsep}{0pt}
\footnotesize
\begin{tabular*}{0.8\textwidth}{@{\extracolsep{\fill}} l cc cc cc c }
\toprule
\multirow{2}{*}{\textbf{Class}} & 
\multicolumn{2}{c}{\textbf{YOLO11}} & 
\multicolumn{2}{c}{\textbf{RF-DETR}} & 
\multicolumn{2}{c}{\textbf{D-FINE}} & 
\multirow{2}{*}{\textbf{Ours}} \\
\cmidrule(lr){2-3} \cmidrule(lr){4-5} \cmidrule(lr){6-7}
 & \textbf{STFT} & \textbf{LS-STFT} & \textbf{STFT} & \textbf{LS-STFT} & \textbf{STFT} & \textbf{LS-STFT} & \\
\midrule
WiFi 20MHz QPSK  & 66.1 & 66.2 & 71.2 & 70.5 & 70.3 & 71.2 & \textbf{71.5} \\
WiFi 20MHz 16QAM & 59.5 & 61.6 & 62.9 & 63.7 & 61.8 & 63.2 & \textbf{63.9} \\
WiFi 20MHz 64QAM & 65.8 & 66.6 & 70.1 & 72.3 & 72.5 & 72.8 & \textbf{72.8} \\
WiFi 40MHz QPSK  & 60.1 & 73.4 & 64.2 & 78.3 & 71.7 & 78.4 & \textbf{78.4} \\
WiFi 40MHz 16QAM & 56.2 & 69.7 & 60.0 & 76.3 & 68.0 & 76.2 & \textbf{76.9} \\
WiFi 40MHz 64QAM & 54.5 & 67.5 & 56.2 & 73.1 & 68.2 & 78.0 & \textbf{78.1} \\
\addlinespace
BLE LE1M         & 49.0 & 63.2 & 53.0 & 75.0 & 65.2 & 79.7 & \textbf{83.0} \\
BLE LE2M         & 43.8 & 59.7 & 48.5 & 71.3 & 57.8 & 78.7 & \textbf{86.8} \\
Zigbee           & 35.7 & 58.7 & 39.1 & 68.1 & 49.5 & 79.8 & \textbf{87.1} \\
LoRa 250kHz      & 31.7 & 55.2 & 35.4 & 69.4 & 46.6 & 77.4 & \textbf{87.1} \\
\addlinespace
SRRC QPSK        & 49.2 & 62.2 & 53.4 & 71.0 & 61.3 & 75.3 & \textbf{76.8} \\
SRRC 16QAM       & 51.4 & 62.3 & 48.9 & 70.1 & 57.6 & 76.5 & \textbf{83.2} \\
AM               & 43.5 & 55.1 & 43.4 & 62.6 & 51.8 & 71.5 & \textbf{83.6} \\
FM               & 37.7 & 56.4 & 41.3 & 56.9 & 46.1 & 61.5 & \textbf{64.2} \\
\bottomrule
\end{tabular*}
\end{table*}

\subsubsection{Model Complexity and Latency Analysis}

To verify that the observed accuracy gains do not simply come from scaling up the network, we report the model size and inference latency of all detectors in Table~\ref{tab:complexity}. 
All models operate on $640 \times 640$ inputs. As shown in the table, the parameter counts are kept within the same order of magnitude (approx. 60M), except for the lightweight RF-DETR which is designed for extreme compression. 
Specifically, ZoomSpec has 60.7M parameters, which is comparable to D-FINE and YOLO11. 
This confirms that the performance superiority of our method stems from the physics-guided architecture rather than significantly increasing model capacity.

Crucially, our analysis highlights the superior accuracy-efficiency trade-off of ZoomSpec. 
Latency is measured with batch size 1 on a single GPU at FP32 precision. 
While the one-stage YOLO11 exhibits the lowest latency, this speed advantage comes at the cost of significant detection failure on narrowband signals. 
ZoomSpec accepts a marginal latency increase to achieve a massive accuracy leap, which is a necessary trade-off for safety-critical monitoring tasks where missing a target is unacceptable.

More importantly, compared to D-FINE, which similarly employs a coarse-to-fine refinement paradigm, ZoomSpec is 24\% faster while achieving higher accuracy. 
This empirically proves that our physics-guided AHLP module-implemented via efficient vectorized DSP operations-is computationally much cheaper than stacking deep learnable attention layers for refinement. With a frame rate of about 60 FPS, ZoomSpec fully satisfies the real-time requirements of low-altitude spectrum sensing systems.

Furthermore, we investigate the scalability of our two-stage architecture under dense signal conditions. A common limitation of cascade frameworks is the risk of linear latency growth with the 
number of detected targets (e.g., sequentially executing AHLP and FRN for 20 concurrent emitters). 
ZoomSpec overcomes this bottleneck via parallelized tensor batching. In our implementation, all candidate proposals from the CPN are stacked into a unified tensor batch, allowing the parallelized AHLP operators and the lightweight FRN to process all candidates concurrently on the GPU. 
Empirical evaluations demonstrate that scaling the number of concurrent signals from 1 to 20 incurs a marginal latency overhead of less than 3\,ms. This confirms that ZoomSpec effectively capitalizes on the inherent sparsity of the radio spectrum-allocating computational resources strictly to active regions-thereby avoiding the computational redundancy of processing the entire wideband noise floor.
\begin{table}[t]
\centering
\caption{Model complexity and latency of compared methods on SpaceNet dataset. 
Latency is measured with batch size 1 on a single GPU.}
\label{tab:complexity}
\renewcommand{\arraystretch}{1.15}
\begin{tabular}{lccc}
\toprule
\textbf{Method} & \textbf{Latency (ms)} & \textbf{Params (M)} & \textbf{Input size} \\
\midrule
YOLO11   & 11.3 & 59.3 & $640 \times 640$ \\
RF-DETR  & 15.1 & 33.7 & $640 \times 640$ \\
D-FINE   & 22.1 & 62.0 & $640 \times 640$ \\
Ours     & 16.8 & 60.7 & $640 \times 640$ \\
\bottomrule
\end{tabular}
\end{table}

\section{Ablation Studies}

\subsection{Effect of Spectral Representation on the CPN}

We first ablate the effect of spectral representation on the Coarse Proposal Net (CPN).  
The CPN is trained with an identical architecture and loss function, but with two different front-end representations: the conventional linear-frequency STFT and the proposed LS-STFT.  
To isolate the quality of coarse bandwidth screening, all modulation families are collapsed into three bandwidth regimes (narrow/mid/wide) plus background, forming a 4-way grading task.

\begin{figure*}[t]
    \centering
    \includegraphics[width=1.3\columnwidth]{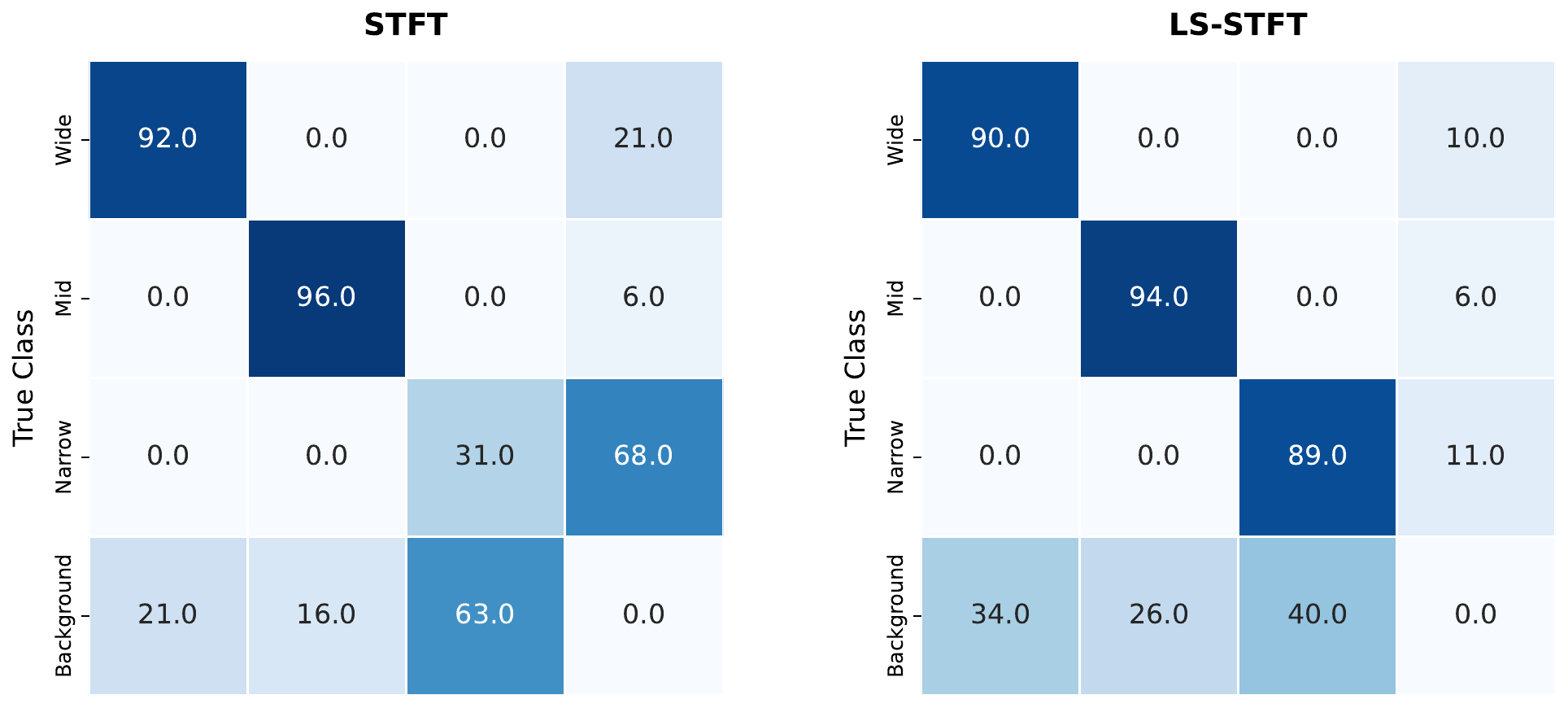}
    \caption {Confusion matrices for the 4-way bandwidth grading task in the CPN under different spectral representations. Standard STFT fails to resolve narrowband emissions, misclassifying 68.0\% of them as background. The proposed LS-STFT significantly enhances narrowband visibility, boosting the recall rate from 31.0\% to 89.0\% while maintaining high accuracy for wideband signals.}
    \label{fig:cpn-confusion}
\end{figure*}

The row-normalized confusion matrices under the two representations are compared in Figure~\ref{fig:cpn-confusion}.
With STFT, mid-band and wide-band emissions are already easy to detect, reaching 92\% and 96\% recall, respectively.  
However, narrow-band signals are severely under-detected: only 31\% of true narrow bands fall into the correct bucket, while 68\% are incorrectly rejected as background.  
This confirms that the linear-frequency STFT fails to resolve narrow occupied bands, causing the system to lose many true emissions at the earliest proposal stage.

With LS-STFT, mid and wide bands remain highly reliable, while narrow-band recall increases dramatically from 31\% to 89\%.  
Although some background samples are still absorbed into the signal buckets, the proposal stage is intentionally recall-oriented; false positives can be eliminated by downstream AHLP purification and the fine detector.  
Overall, these results indicate that the log-space frequency mapping effectively allocates more resolution to narrow bands, sharpens spectral edges, and enables the CPN to preserve most true candidates for subsequent stages.

\begin{table*}[!t]
\centering
\caption{Ablation study on LS-STFT, CPN, AHLP, and FRN variants. 
mAP is evaluated as mAP@0.5:0.95.}
\label{tab:ablation}
\renewcommand{\arraystretch}{1.25}
\setlength{\tabcolsep}{6pt}
\begin{tabular}{lcccccc}
\toprule
\textbf{Variant} & \textbf{LS-STFT} & \textbf{CPN} & \textbf{AHLP} & \textbf{FRN} & \textbf{Fusion Mode} & \textbf{mAP@0.5:0.95} \\
\midrule
Baseline with STFT                    & \xmark & \cmark & \xmark & \cmark & full        & 58.8 \\
+ LS-STFT                             & \cmark & \cmark & \xmark & \cmark & full        & 72.3 \\
\midrule
\multicolumn{7}{l}{\textit{Fusion ablations within FRN}} \\
I/Q only                        & \cmark & \cmark & \cmark & \cmark & I/Q only    & 66.5 \\
FFT only                        & \cmark & \cmark & \cmark & \cmark & FFT only    & 43.7 \\
w/o cross-domain fusion         & \cmark & \cmark & \cmark & \cmark & no fusion   & 76.7 \\
\midrule
Full model                            & \cmark & \cmark & \cmark & \cmark & full        & \textbf{78.1} \\
\bottomrule
\end{tabular}
\end{table*}

\subsection{Effect of AHLP and Dual-Domain Fusion}

Table~\ref{tab:ablation} summarizes the ablation results for LS-STFT, CPN, AHLP, and the dual-domain FRN.  
Switching the front-end from STFT to LS-STFT improves mAP@0.5:0.95 from 58.8 to 72.3, validating the importance of constant relative resolution and better narrow-band visibility.  
Building on this stronger representation, AHLP further improves mAP from 72.3 to 78.1 in the full system, contributing a substantial gain of 5.8 points.  
This improvement comes primarily from removing cross-band spectral leakage and stabilizing the bandwidth geometry before entering the fine detector.

We additionally evaluate three variants of the FRN to isolate the importance of dual-domain fusion.  
Using only the I/Q branch degrades mAP to 66.5, while relying solely on the FFT domain collapses performance to 43.7, indicating that neither domain alone is sufficient.  
Removing cross-domain fusion but keeping both branches active yields 76.7 mAP, still below the full 78.1 mAP achieved with dual-domain fusion.  
These comparisons demonstrate that the I/Q and LS-STFT features are complementary and that fusion is essential for fully exploiting their strengths.

\section{Conclusion}
This paper introduced ZoomSpec, a dual-domain two-stage framework for wideband spectrum sensing. By combining LS-STFT, coarse proposals, 
physics-guided AHLP purification, and cross-domain fusion, the system achieves markedly improved localization and recognition accuracy. Evaluations on the official SpaceNet dataset show consistent gains 
over STFT- and LS-STFT–based detectors and surpass the top reported challenge result. The findings highlight the effectiveness of integrating physical priors with learned representations for robust spectrum sensing.

\bibliographystyle{IEEEtran}
\bibliography{ref}

\end{document}